\documentclass[times, review, 10pt]{elsarticle}

\usepackage{amssymb}
\usepackage{amsmath}
\usepackage{pgfplots}
\pgfplotsset{compat=1.15}
\usepackage{amsmath}
\usepackage{amssymb}
\usepackage{multirow}
\usepackage[numbers]{natbib}
\newcommand\Tstrut{\rule{0pt}{2.3ex}}

\newcommand*{\img}[1]{
    \raisebox{-.02\baselineskip}{
        \includegraphics[
        height=\baselineskip,
        width=\baselineskip,
        keepaspectratio,
        ]{#1}
    }
}

\begin{document}

\begin{frontmatter}

\title{Learning to Reason and Navigate: Parameter Efficient Action Planning with Large Language Models}

\author[adelaide]{\vspace{-3mm}Bahram Mohammadi}

\author[adelaide]{Ehsan Abbasnejad}
\author[sydney]{Yuankai Qi}
\author[adelaide]{\\Qi Wu}
\author[adelaide]{Anton Van Den Hengel}
\author[adelaide]{Javen Qinfeng Shi}

\date{}

\makeatletter
\def\ps@pprintTitle{
 \let\@oddhead\@empty
 \let\@evenhead\@empty
 \let\@oddfoot\@empty
 \let\@evenfoot\@empty}
\makeatother

\affiliation[adelaide]{organization={Australian Institute for Machine Learning},
            addressline={University of Adelaide}, 
            city={Adelaide},
            state={SA},
            country={Australia}}

\affiliation[sydney]{organization={Center for Applied Artificial Intelligence},
            addressline={Macquarie University}, 
            city={Sydney},
            state={NSW},
            country={Australia}}

\begin{abstract}
The remote embodied referring expression (REVERIE) task requires an agent to navigate through complex indoor environments and localize a remote object specified by high-level instructions, such as ``\textit{bring me a spoon}'', without pre-exploration. Hence, an efficient navigation plan is essential for the final success. This paper proposes a novel parameter-efficient action planner using large language models (PEAP-LLM) to generate a single-step instruction at each location. The proposed model consists of two modules, LLM goal planner (LGP) and LoRA action planner (LAP). Initially, LGP extracts the goal-oriented plan from REVERIE instructions, including the target object and room. Then, LAP generates a single-step instruction with the goal-oriented plan, high-level instruction, and current visual observation as input. PEAP-LLM enables the embodied agent to interact with LAP as the path planner on the fly. A simple direct application of LLMs hardly achieves good performance. Also, existing hard-prompt-based methods are error-prone in complicated scenarios and need human intervention. To address these issues and prevent the LLM from generating hallucinations and biased information, we propose a novel two-stage method for fine-tuning the LLM, consisting of supervised fine-tuning (STF) and direct preference optimization (DPO). SFT improves the quality of generated instructions, while DPO utilizes environmental feedback. Experimental results show the superiority of our proposed model on REVERIE compared to the previous state-of-the-art.
\end{abstract}


\end{frontmatter}

\makeatletter
\def\ps@pprintTitle{
 \let\@oddhead\@empty
 \let\@evenhead\@empty
 \let\@oddfoot\@empty
 \let\@evenfoot\@empty}
\makeatother

\section{Introduction}
\label{sec:introduction}
Vision-and-language navigation (VLN)~\cite{anderson_2018_vln} is a complex task involving different areas such as computer vision, natural language processing, and navigation. In recent years, VLN has drawn significant attention from these research communities~\cite{an_2023_bevbert, li_2023_kerm, lin_2022_adapt, wang_2021_ssm, hong_2020_language, deng_2020_evolving, guhur_2021_airbert, liu_2021_mixup, li_2019_robust}. In general, VLN requires an agent to navigate through the 3D-simulated environments by following human instructions. Given the substantial potential of VLN in real-world applications, various VLN tasks have been introduced. Room-to-room (R2R)~\cite{anderson_2018_vln} and room-across-room (RxR)~\cite{ku_2020_rxr} necessitate an agent to reach a pre-specified location by following fine-grained step-by-step instructions. Remote embodied visual referring expression in real indoor environments (REVERIE)~\cite{qi_2020_reverie} and scenario-oriented object navigation (SOON)~\cite{zhu_2021_soon} introduce a goal-oriented VLN task in which the embodied agent needs to find a remote object that is not visible in the starting location. In contrast to most VLN tasks, which include detailed instructions, REVERIE contains high-level and short instructions to be more practical and closer to human preferences. Thus, providing the agent with a proper navigation plan is essential for efficient exploration.

Large language models (LLMs) have recently demonstrated significant promise in action planning. However, they are not well-explored in the REVERIE task, which contains complex environments with multi-layer buildings and a wide range of objects. NavGPT~\cite{zhou_2024_navgpt} and MapGPT~\cite{chen_2024_mapgpt} investigate the LLM capabilities for zero-shot navigation, however, there is a large gap between their performance and supervised methods. To show the effectiveness of language for navigation tasks,  NavCoT~\cite{lin_2024_navcot} and LangNav~\cite{pan_2023_langnav} finetune an LLM on VLN data. MiC~\cite{qiao_2023_march} provides the agent with additional language instructions using in-context learning~\cite{brown_2020_language}, which may encounter challenges when addressing specialized domains. Although flexible, this approach highly depends on the quality of input prompts. Moreover, inadequate or biased training data results in sub-optimal performance.

\begin{figure}[!t]
\begin{center}
\includegraphics[width=0.99\linewidth]{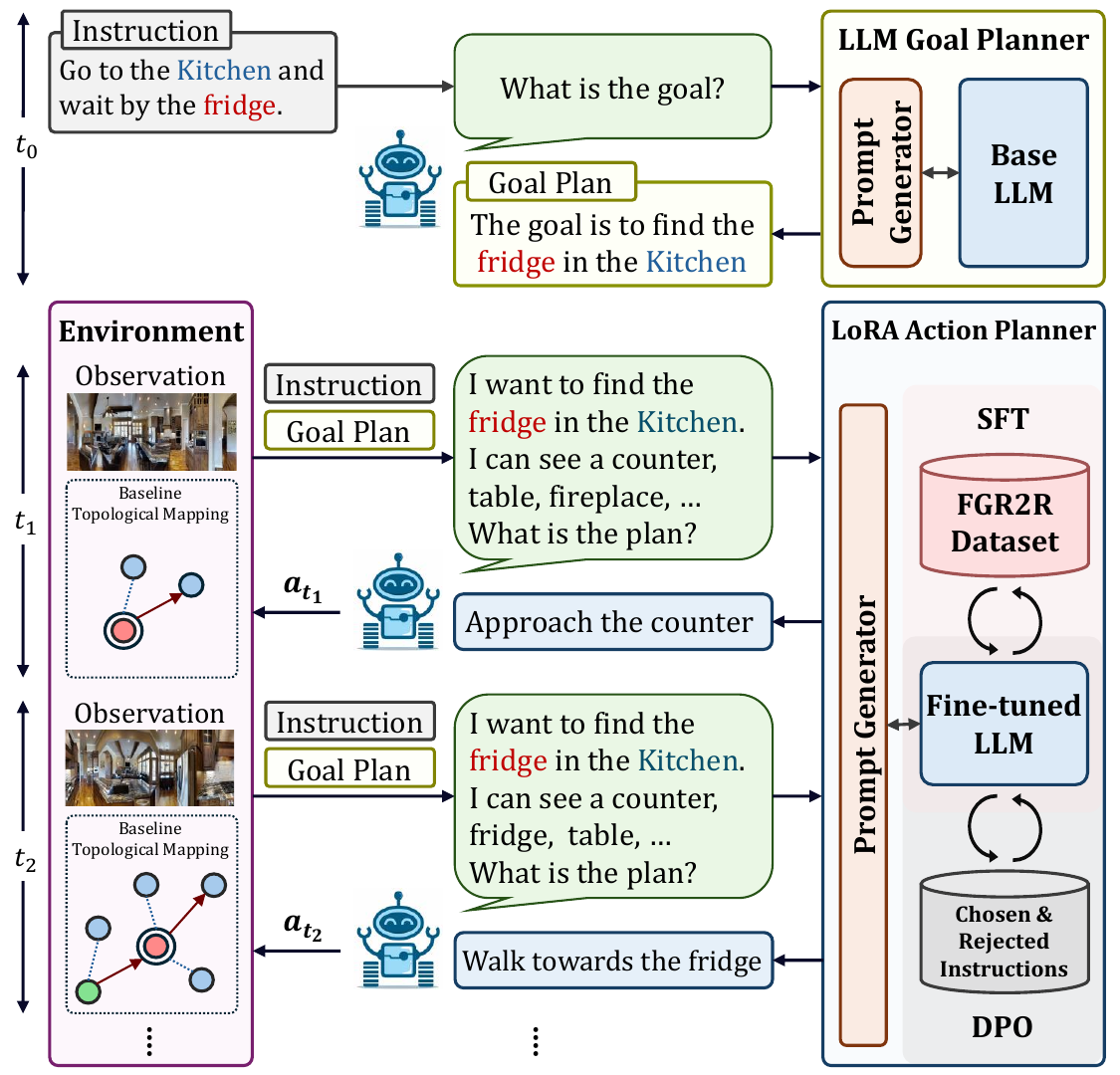}
\vspace{-5mm}
\end{center}
   \caption{At time step $t_0$, the LLM goal planner specifies the target object and room using the base LLM. Then, the LoRA action planner utilizes the fine-tuned LLM to generate a single-step instruction with the goal-oriented plan, high-level instruction, and visual observation as input. The base LLM is fine-tuned in two SFT and DPO settings using the FGR2R dataset and environmental feedback, respectively.
   }
\label{fig::teaser}
\end{figure}

To tackle the above-mentioned issues and by exploiting the reasoning ability of LLMs in navigation, we propose a novel parameter-efficient action planner using LLMs (PEAP-LLM) consisting of two modules, LLM goal planner (LGP) and LoRA action planner (LAP) as Fig.~\ref{fig::teaser} shows. At first, LGP queries the base LLM to retrieve the target object and the room where it is located. Then, LAP is asked to generate a single-step instruction with a prompt as input, which contains the goal-oriented plan, high-level instruction, and the most informative visible objects in the panoramic view. Our model enables the embodied agent to interact with the LAP on the fly as the path planner. Eventually, the next action is predicted by the baseline policy model, \textit{i.e.}, HM3D-DUET~\cite{chen_2022_hm3d}, with respect to both high-level REVERIE instruction and the generated single-step action plan.

Based on our observations, LLMs are helpful but not strong enough to generate efficient action plans. For example, the base LLM tends to use certain objects in generated sub-instructions or mention objects that are not present in the scene. These problems indicate generating biased information and hallucinations, respectively, leading to an incorrect action prediction. Furthermore, designing an accurate prompt to query the LLM requires human intervention, which is susceptible to error. To mitigate these issues, we propose a novel approach to fine-tune the LLM on the REVERIE downstream task in two steps: (1) supervised fine-tuning (SFT) using parameter-efficient fine-tuning (PEFT) techniques, including prefix tuning~\cite{li_2021_prefix} and low-rank adaptation (LoRA)~\cite{hu_2021_lora}, and (2) direct preference optimization (DPO)~\cite{rafailov_2024_direct} to incorporate the environmental feedback into the fine-tuning process. The major obstacle to doing so is utilizing or generating the appropriate dataset. For PEFT, we need a set of ground-truth sub-instructions and their corresponding prompts. The ground-truth trajectories in REVERIE are identical to R2R, therefore, we use the decomposed step-by-step instructions
of R2R provided by fine-grained R2R (FGR2R)~\cite{hong_2020_sub}. For DPO, it is essential to construct the preference dataset by distinguishing between the preferred and dispreferred responses according to the feedback from the action prediction process of the policy model. We use single-step action prediction (SAP) task~\cite{chen_2021_history} to evaluate whether a sub-instruction is preferred or dispreferred.

Experimental results on REVERIE show the effectiveness and generalization of our proposed method, which outperforms the previous state-of-the-art approaches. The main contributions of this work are summarized as follows:

\begin{enumerate}
    \item 
    We propose a novel parameter-efficient action planner (PEAP-LLM)  that enables the agent to interact with the LLM on the fly as the path planner to obtain a single-step instruction in each location.
        
    \item
    We propose two modules: LGP to extract target objects and rooms from REVERIE instructions and LAP to generate a single-step action plan at each time step.

    \item
    We propose a novel approach to collect appropriate datasets and fine-tune the LLM on the REVERIE downstream task to enhance the navigation and object grounding ability of the agent.
 
    \item
    The experiments are conducted on the REVERIE benchmark. Our model improves the baseline performance by $4.00$\% on SPL and $3.20$\% on RGSPL for the validation unseen split and $2.10$\% on SPL and $2.20$\% on RGSPL for the test unseen split and achieves new state-of-the-art.
\end{enumerate}

\section{Related Work}
\label{sec:related_work}
\noindent \textbf{Vision-and-Language Navigation.}
Vision-and-language navigation (VLN) has garnered growing interest in recent years, which has led to introducing a wide range of VLN tasks~\cite{anderson_2018_vln, ku_2020_rxr, qi_2020_reverie, zhu_2021_soon, thomason_2020_vision, chen_2019_touchdown,askvln,countervln,goldseeker}. The first task, \textit{i.e.}, R2R, is proposed by Anderson \textit{et al.}~\cite{anderson_2018_vln}, which is about an agent navigating through real-world indoor environments to reach the target location from a starting point by strictly following step-by-step instructions. Afterward, Ku \textit{et al.}~\cite{ku_2020_rxr} propose RxR, a multilingual dataset (English, Hindi, and Telugu) with more paths and instructions. The main focus of R2R and RxR is navigation, while Qi \textit{et al.}~\cite{qi_2020_reverie} introduce REVERIE, requiring the agent to explore the environment and localize a remote object regarding concise high-level instructions. Similar to REVERIE, Zhu \textit{et al.}~\cite{zhu_2021_soon} propose SOON, a goal-oriented task but with longer and more detailed instructions.

\vspace{1mm}
\noindent \textbf{Incorporating LLMs into Navigation.}
In light of the benefits LLMs bring to various tasks, they have been utilized in recent works for navigation in the area of embodied artificial intelligence. SayCan~\cite{ahn_2022_can} converts high-level instructions into a series of low-level actions to predict the next action. \cite{huang_2022_language} employs frozen LLMs to predict actions for the embodied agent through in-context learning. Inner Monologue~\cite{huang_2022_inner} utilizes LLMs for action planning without further training by leveraging environmental feedback such as success detection, scene description, and human interaction. NavGPT~\cite{zhou_2024_navgpt} shows the potential capability of LLMs for navigational reasoning with a complicated prompting process. DiscussNav~\cite{long_2023_discussnav} introduces a zero-shot VLN framework using multi-expert discussion. MapGPT~\cite{chen_2024_mapgpt} proposes a topological map in the linguistic form to enhance global exploration. NavCoT~\cite{lin_2024_navcot} and LangNav~\cite{pan_2023_langnav} fine-tune an LLM on VLN data to investigate the effectiveness of language to perform navigation. MiC~\cite{qiao_2023_march} proposes an environment-aware LLM as a path planner through on-the-fly interactive communication between the agent and the LLM, leveraging in-context learning.

\vspace{1mm}
\noindent \textbf{Parameter-Efficient Fine-Tuning.} 
PEFT techniques such as prefix tuning~\cite{li_2021_prefix} and LoRA~\cite{hu_2021_lora} provide an efficient remedy by diminishing the number of fine-tuning parameters and memory usage while attaining performance levels comparable to fully fine-tuning. Prefix-tuning offers a lightweight alternative for fine-tuning natural language generation models, which maintains language model parameters frozen, but optimizes a small set of task-specific vectors. LoRA freezes the pre-trained model weights and injects trainable rank decomposition matrices into each layer of the Transformer architecture.

\begin{figure*}[!t]
\begin{center}
\includegraphics[width=0.99\linewidth]{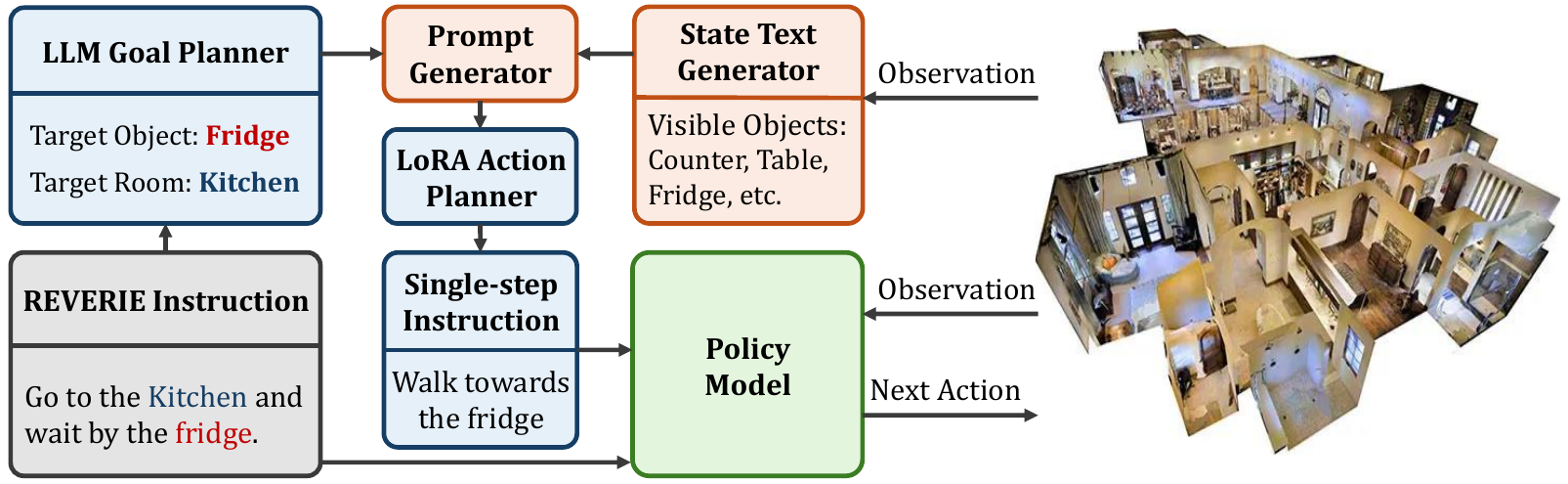}
\end{center}
   \caption{Outline of our PEAP-LLM model. At first, we perform the goal-oriented planning task using the LGP module. Then, LAP takes the goal-oriented plan, high-level instruction, and the current visual observation as input to offer a single-step navigation plan. Finally, the policy model predicts the next action and the process is repeated until the episode ends by predicting the stop action or exceeding the maximum number of steps.}
\label{fig::outline}
\end{figure*}

\section{Proposed Method}
\label{sec:proposed_method}
In this section, we initially elaborate on the problem formulation and the outline of our proposed model. Afterward, we comprehensively explain PEAP-LLM, encompassing the LGP and LAP modules. Also, we present our approach for fine-tuning the LLM on the REVERIE downstream task in SFT and DPO settings.

\vspace{1mm}
\noindent\textbf{Problem Formulation.} 
REVERIE benchmark contains short and concise instructions denoted as $I = \{w_i\}^{L}_{i=1}$, where $w_{i}$ is the $i^{th}$ word and $L$ is the length of the instruction. At each time step, the agent receives a panoramic view of its surroundings $\mathcal{V}_{t} = \{v_{t, i}\}^{36}_{i=1}$. The agent infers an action $a_{t}$ to transfer from state $s_{t}$ to state $s_{t+1}$ only through the navigable directions $\mathcal{N}(\mathcal{V}_{t}) = \{v_{t, i}\}^{K}_{i=1}$, where $\mathcal{N}(\mathcal{V}_{t}) \subseteq \mathcal{V}_{t}$. Each state includes a triplet $\{v_{t, i}, \theta_{t, i}, \psi_{t, i}\}$, where $v_{t, i}$ is the viewpoint image, and $\{\theta_{t, i}, \psi_{t, i}\}$ are angels of heading and elevation, respectively, to determine the orientation of the image regarding the agent position. The episode ends when the agent identifies the position of the target object within the panoramic view.

\vspace{1mm}
\noindent\textbf{Method Overview.}
This paper aims to enhance the vision-text alignment as well as the navigation and object grounding ability of the agent by generating complementary single-step instructions at each location. As illustrated in Fig.~\ref{fig::outline}, our method consists of two modules, the LLM goal planner (LGP) and the LoRA action planner (LAP). To improve the quality of the generated sub-instructions, it is necessary to extract the goal of REVERIE instructions, which is in the form of ``\textit{finding the object X in the room Y}''. LGP extracts object X and room Y from each high-level instruction, exploiting the extensive commonsense knowledge implied in the base LLM.

LGP works regardless of the visual input, while the LAP module needs dynamic observations, such as visible objects in the scene, to generate more effective sub-instructions. We filter out less relevant objects and choose the top 5 most informative ones according to the goal of the REVERIE instruction and the visual perception. The goal-oriented plan obtained by LGP, visible objects in the scene, and the high-level instruction are leveraged to query the LAP module for generating single-step instructions. Contrary to LGP, LAP utilizes the fine-tuned LLM.

\subsection{Baseline Agent}
\label{baseline_agent}
In this work, the main baseline is HM3D-DUET~\cite{chen_2022_hm3d}, which follows the architecture of DUET~\cite{chen_2022_duet}. The core concept of the DUET model involves constructing a topological map on the fly, enabling the agent to devise an effective long-term plan across all navigable nodes in the graph rather than adjacent nodes. The model comprises two components: topological mapping and global action planning. The former component is responsible for adding the new visited location and updating node representations. The latter component predicts the next viewpoint in the graph or stop action. The global action planning component consists of two encoders: fine- and coarse-scale encoders. The fine-scale encoder makes local actions based on fine-grained visual representation, while the coarse-scale encoder predicts the next action over all observed nodes during the navigation. The dynamic fusion of both encoders leads to the best performance in navigation and object grounding ability of the agent.

\subsubsection{Fine-scale Encoder}
This module concentrates on local observations at each node of the topological map. The inputs of this module are the contextual word representation $\hat{\mathcal{W}}_{t}$ and the visual representation of the current node which includes the concatenation of encoded embeddings of image and object features, denoted as $\mathcal{R}_{t}$ and $\mathcal{O}_{t}$, respectively. In DUET, two types of location embeddings are added to the visual representation: the relative location of the current node to the starting point and the relative position of the adjacent nodes to the current node. Finally, we have the visual token $[r_{0}; \mathcal{R}_{t}; \mathcal{O}_{t}]$, where $r_{0}$ is the stop action. To model the relations between vision and language and obtain the embeddings of visual tokens, a multi-layer cross-modal transformer~\cite{tan_2019_lxmert} is utilized. 

We need to specify two navigation scores in the local action space $a_t$ along with object scores for the REVERIE task to ground the target object at the final time step. The navigation score of the $v_{i}$ are predicted based on:

\begin{equation}
\label{eq:fine_scale_score}
    \begin{split}
        s^{f}_{i} = \text{FFN}(\hat{v_{i}})
    \end{split}
\end{equation}

\noindent where $\hat{v_{i}}$ is the output embedding of $v_{i}$ which is $[r_{0}; \mathcal{R}_{t}; \mathcal{O}_{t}]$ and FFN demonstrates a two-layer feed-forward neural network. Similarly, $s^{o}_{i}$ is calculated.

\subsubsection{Coarse-scale Encoder}
There are two types of encoding added to the visual representation of each node in DUET: (1) location encoding which is the relative distance and direction of the node regarding the current node, and (2) navigation step encoding resulting in encoding of the visited nodes differently compared to the unvisited nodes to improve visual-textual alignment. To obtain the word and node embedding, DUET introduces graph-aware self-attention (GASA) that slightly differs from the standard attention in transformers. According to Eq.~\ref{eq:gasa}, the structure of the topological map is taken into account to model the environment layout and relations between nodes and instructions.

\begin{equation}
    \begin{split}
        \text{GASA} = \text{Softmax} & \left( \frac{XW_{q}(XW_{k})^{T}}{\sqrt{d}} + M \right) XW_{v} \\
        & M = EW_{e} + b_{e} \\
    \end{split}
    \label{eq:gasa}
\end{equation}

\noindent where $X$ shows node representations, $E$ denotes the distance matrix, and $W_{e}, b_{e}$ are two learnable parameters. The navigation score for each node $V_i$ in the graph is predicted as follows:

\begin{equation}
    \begin{split}
        s^{c}_{i} = \text{FFN}(\hat{v}_i)
    \end{split}
    \label{eq:score}
\end{equation}

\noindent where $\hat{v}_i$ is the output embedding of node $V_i$ and FFN denotes a two-layer feed-forward network. $s^{c}_{0}$ demonstrates the stop score.

\subsubsection{Action Reasoning}
The predicted scores are obtained through the dynamic fusion of fine- and coarse-scale encoders. To ensure consistency between the two encoders, local action scores are converted into the global action space. The converted local action scores are obtained as follows:

\begin{equation}
    s^{f^{\prime}}_{i} = 
    \begin{cases}
        s_{\text{back,}} & \text{if $V_{i} \in \mathcal{V}_{t} - \mathcal{N}(V_{t})$} \\
        s^{f}_{i} & \text{otherwise.}
    \end{cases}
    \label{eq:convered_global_score}
\end{equation}

\vspace{2mm}

To predict a scalar for fusion $\sigma_{t}$, $\hat{v}_{0}$ from coarse-scale encoder is concatenated to $\hat{r}_{0}$ from fine-scale encoder, \textit{i.e.}, $\sigma_{t} = \text{Sigmoid} (\text{FFN}([\hat{v}_{0}; \hat{r}_{0}])) $. The final navigation score for $V_i$ is:

\begin{equation}
    \begin{split}
        s_i = \sigma_t s^{c}_{i} + (1-\sigma_t) s^{f^\prime}_{i}
    \end{split}
    \label{eq:fusion_scalar}
\end{equation}

\subsubsection{Baseline Policy Model Optimization}
\label{instruction_generator::interactive_prompting}
Alongside the behavior cloning tasks, including single-step action prediction (SAP) \cite{chen_2021_history} and object grounding (OG)~\cite{lin_2021_scene}, the baseline model is also pre-trained on two additional auxiliary proxy tasks, masked language modeling (MLM)~\cite{devlin_2019_bert}, masked view classification (MVC)~\cite{lu_2019_vilbert}. Optimization of the policy model is performed by the supervision provided by a pseudo-interactive demonstrator instead of behavioral cloning, similar to the DAgger algorithm~\cite{reduction_2011_ross}. Since the environment graph is available during the training phase, the agent selects the next location with the overall shortest distance to the destination.

\subsection{LLM Goal Planner}
\label{method::llm_goal_planner}
To improve the generated single-step instructions, it is crucial to identify the target object and room associated with each high-level instruction. Since extracting goal-oriented plans regardless of visual observation is not complicated, this module utilizes the base LLM. However, acquiring this information directly from instructions without intermediate reasoning steps is prone to errors. Inspired by chain-of-thought (CoT) prompting~\cite{wei_2022_chain}, we decompose the entire process of goal-oriented planning into three sequential phases conducted by the base LLM: (1) extracting the goal of the high-level instruction, (2) identifying the target object regarding the goal, and (3) specifying the target room based on the high-level instruction and the usual whereabouts of the target object. For instance, considering the REVERIE instruction ``\textit{Bring me the trash can from the bathroom}'', the LLM is first asked to determine the main goal ``\textit{Bring trash can}''. Then, ``\textit{trash can}'' is identified as the target object. Finally, ``\textit{bathroom}'' is specified as the target room regarding the instruction and target object. 

Based on our observations, identifying the target room can be challenging for the base LLM under certain circumstances. To be more precise, we utilize a list of rooms sourced from the REVERIE dataset. Selecting the target room from this list brings two benefits: (1) recognizing rooms that are not very common in a house like "\textit{office room}", and (2) identifying the target room, which is not explicitly mentioned in the instruction. To reduce training time, LGP is performed once before starting the task. Note that considering a list of rooms for determining the target room does not harm the generalization ability of the agent. This is because, in REVERIE, similar to many other VLN benchmarks, the agent is required to explore indoor environments, which contain a definite number of room types, to localize the target object. REVERIE covers almost all room types in the real world.

\subsection{LoRA Action Planner}
\label{method::lora_action_planner}
LAP generates a single-step instruction at each time to build a strong correspondence between the visual and textual data and improve the traceability of agents through the completion of instructions. Contrary to the LGP module, the LAP, as its name suggests, exploits the fine-tuned version of the LLM. Visible objects along the path are leveraged as visual clues in enhancing navigation and object grounding performance. Therefore, LAP incorporates dynamic information into the action plan generation in addition to the high-level instruction and goal-oriented plan. In the following sections, we explain the object retrieval process and the two-stage approach for fine-tuning the LLM in SFT and DPO settings. Unlike LGP, which is called once before starting the task, LAP runs at each location along the path until the agent stops or exceeds the maximum number of steps.

\vspace{1mm}
\noindent \textbf{Object Retrieval.}
\label{method::object_retrieval}
In this work, we use open-vocabulary diffusion-based panoptic segmentation (ODISE)~\cite{xu_2023_odise} to extract visible objects. Due to the high time complexity of identifying object labels for all viewpoint images at time step $t$, $\{v_{t, i}\}^{36}_{i=1}$, ODISE is executed for the panoramic view $p_t$. The top 5 most pertinent objects are detected from object sets $\mathcal{O}_{p_{t, i}}$ using contrastive language-image pre-training (CLIP)~\cite{radford_2021_Learning} model. This model consists of CLIP-I and CLIP-T for encoding images and texts into a joint embedding space. All images and object labels are encoded by CLIP-I and CLIP-T, respectively. We retrieve the most relevant objects regarding their distance score with average encoded viewpoint images, which are calculated by the cosine similarity as:

\begin{equation}
    s(\mathcal{V}_{t}, o_{t, i}) =
    \frac{\overline{e(\mathcal{V}_t)}~\cdot~e(o_{t, i})}{\vert\vert \overline{e(\mathcal{V}_t)} \vert\vert ~ \vert\vert e(o_{t, i}) \vert\vert}
\end{equation}

\noindent
where $e(.)$ is the encoding function. A greater score for an object label indicates it is more relevant with respect to the surrounding environment.

\subsection{Prompting Template}
\label{sec:prompting_template}
In this section, we provide the prompting templates for LGP and LAP. Also, we elucidate how we prompt in particular the Llama 2~\cite{touvron_2023_llama} model.

\begin{figure*}[!t]
\begin{center}
\includegraphics[width=0.99\linewidth]{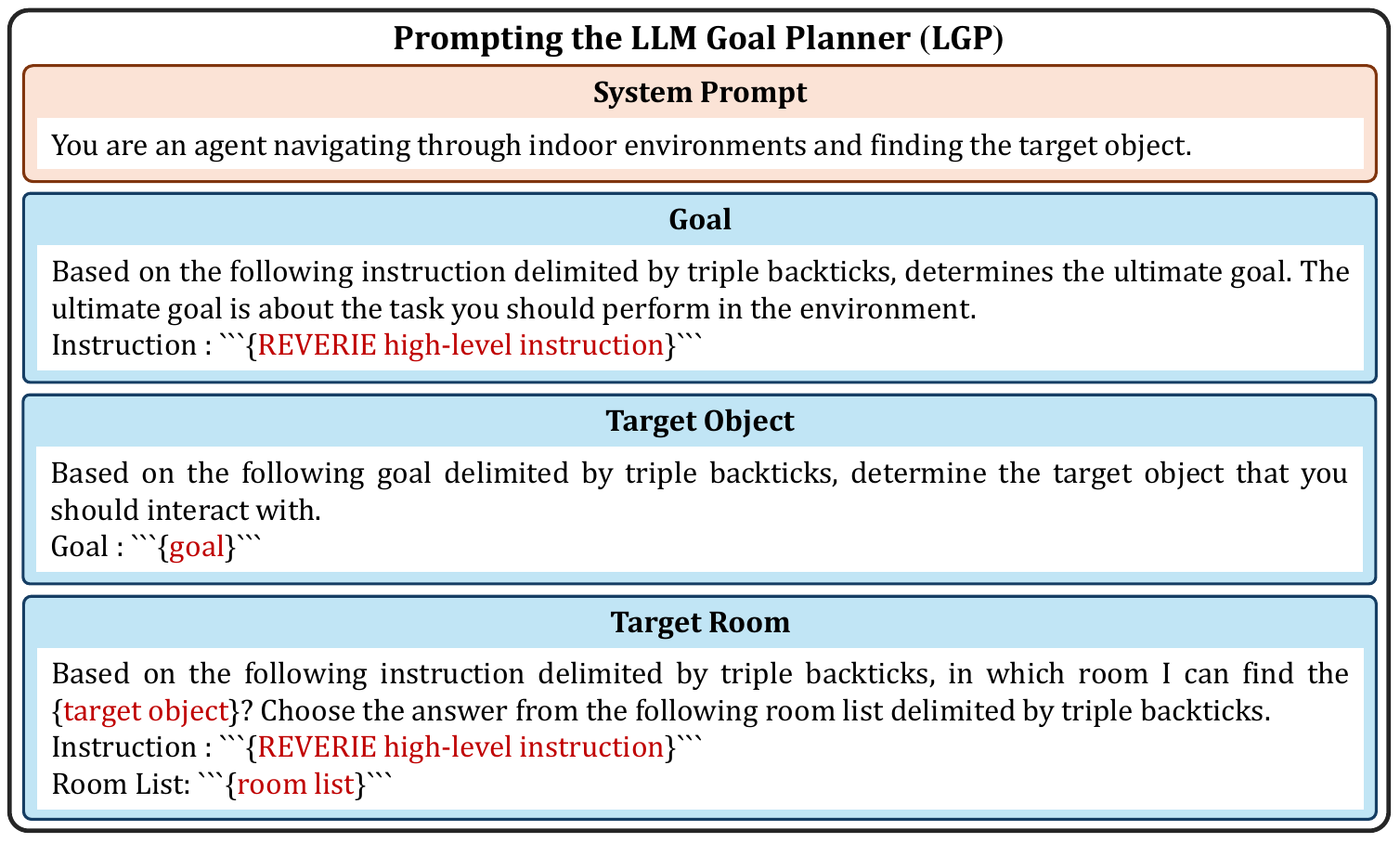}
\vspace{-4mm}
\end{center}
   \caption{Example of prompting template for LLM goal planner. The \textcolor{red}{red} denotes the input parameters.}
\label{fig::lgp_prompting}
\end{figure*}

\subsubsection{LGP Prompting}
As discussed in Section~\ref{method::llm_goal_planner} and according to Fig.~\ref{fig::lgp_prompting}, the target object and room are obtained using the LGP module. At first, we define the system prompt to inform the model about the general task. Afterward, object recognition and object localization are performed in three sequential steps: (1) extract the main goal of the REVERIE high-level instruction, (2) retrieve the target object according to the goal retrieved from the previous step, and (3) find the target room based on the usual location of the target object and the REVERIE instruction. To obtain a more precise response from the base LLM, the target room is selected from a list sourced from REVERIE. This benchmark covers almost all room types in the real world including ``\textit{bathroom, bedroom, closet, dining room, entryway, family room, garage, hallway, library, laundry room, kitchen, living room, meeting room, lounge, office room, porch, rec room, stairs, toilet, utility room, gym, outdoor, workout room, bar, classroom, dining booth}, and \textit{spa}''.

\begin{figure*}[!t]
\begin{center}
\includegraphics[width=0.99\linewidth]{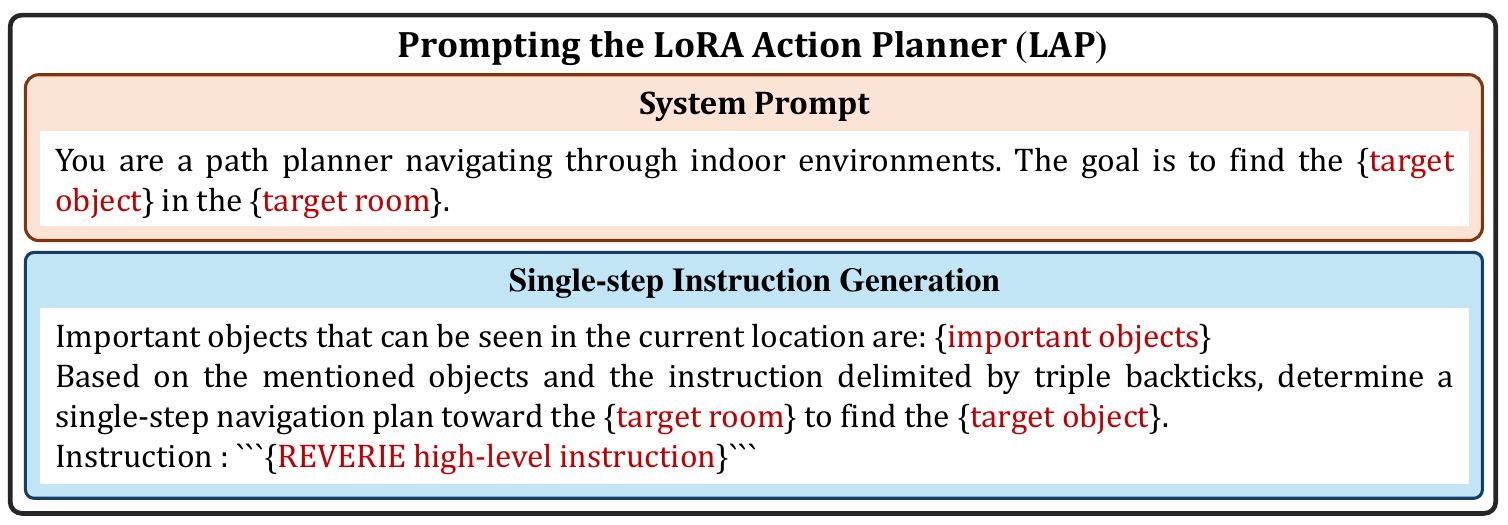}
\vspace{-4mm}
\end{center}
   \caption{Example of prompting template for LoRA action planner. The \textcolor{red}{red} denotes the input parameters.}
\label{fig::lap_prompting}
\end{figure*}

\subsection{LAP Prompting}
Similar to the prompt for LGP, we first consider a system prompt to describe the general task of the agent, which is finding object \textit{X} in room \textit{Y} for generating a single-step action plan as illustrated in Fig.~\ref{fig::lap_prompting}. In this module, objects play a crucial role in generating more accurate and informative single-step instructions. Hence, LAP outputs the final action plan with the goal-oriented plan, REVERIE high-level instruction, and visible objects as input.

\subsection{Prompting the Llama 2 Model}
The prompt template for a single message instance using the Llama 2 model is:
\newline
\newline
{\fontfamily{cmtt}\selectfont
\noindent
<s>[INST] <<SYS>> \\
\{\{ system\_prompt \}\} \\
<</SYS>> \\
\{\{ user\_message \}\} [/INST]
}
\newline

\noindent while for multiple messages, the above template is followed by:
\newline
\newline
{\fontfamily{cmtt}\selectfont
\noindent
\{\{ model\_answer\_1 \}\} </s> \\
<s>[INST] \{\{ user\_message\_2 \}\} [/INST]
}
\newline

Open-access models offer the opportunity to exert full control over the {\fontfamily{cmtt}\selectfont system} prompt within chat applications, which can be considered crucial to defining the behavior of the chat assistant.

\subsection{LLM Fine-tuning}
\label{method::lora_fine_tuning}
LLMs often tend to generate responses even when their knowledge of a specific topic is not sufficient. This can lead to producing misleading or completely incorrect information. This behavior, where the model aims to provide convincing but potentially false answers, is referred to as hallucination. To minimize generating hallucinations, mitigate biases present in LLMs, and reduce the impact of human intervention in designing an accurate prompt, we fine-tune the LLM utilizing SFT and DPO. The main challenge of fine-tuning the LLM lies in constructing a proper dataset to make it a VLN specialist, which is explained in the following sections.

\begin{figure*}[!t]
\begin{center}
\includegraphics[width=0.99\linewidth]{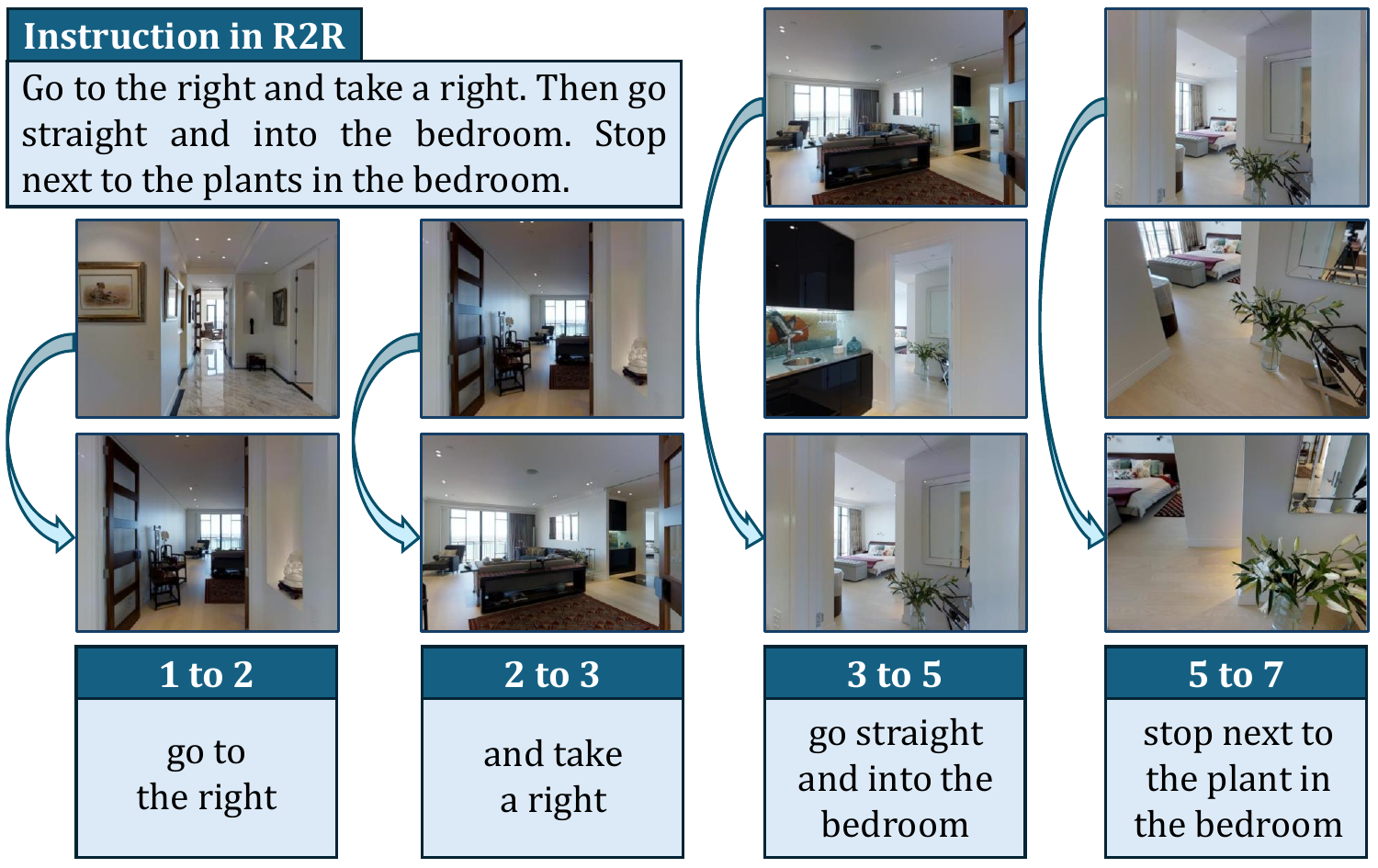}
\end{center}
   \caption{The process of decomposing the entire R2R instructions into non-overlapping sub-instructions based on the chunk view specifying which sub-instruction corresponds to which location. $x$ to $y$ shows the determined sub-instruction is suitable for going to viewpoint $y$ from viewpoint $x$.}
\label{fig::fgr2r}
\end{figure*}

\subsubsection{Supervised Fine-tuning.}
To fine-tune the LLM in the supervised setting, it is necessary to construct a dataset containing a set of ground-truth instructions and the corresponding prompts in each viewpoint. In this setting, we aim to extract the ground-truth sub-instructions from the FGR2R~\cite{hong_2020_sub} dataset. This idea is reasonable because although the tasks are different, the ground-truth trajectory is shared between the REVERIE and R2R benchmarks. As shown in Fig.~\ref{fig::fgr2r}, FGR2R decomposes the entire R2R instructions into several non-overlapping sub-instructions. In this example, the agent is required to pass through seven viewpoints to reach the target location by strictly following the instructions retrieved from the R2R benchmark, and divided into four sub-instructions. Since we aim to generate single-step instructions for all locations, each viewpoint is required to map onto a generated action plan. The FGR2R dataset includes the chunk view element determining which sub-instruction brings the agent to the next navigable direction through the ground-truth path. In this work, we employ two PEFT methods, prefix tuning and low-rank adaptation (LoRA). To achieve the final results in this paper, LoRA is utilized owing to its considerable advantages, such as faster training and adaptation, as well as feasibility for smaller hardware. In this setting, we fine-tune the base LLM to prepare the base LAP for the next phase, \textit{i.e.}, fine-tuning in the DPO setting, which results in the final model for action prediction.

\begin{figure*}[!t]
\begin{center}
\includegraphics[width=0.99\linewidth]{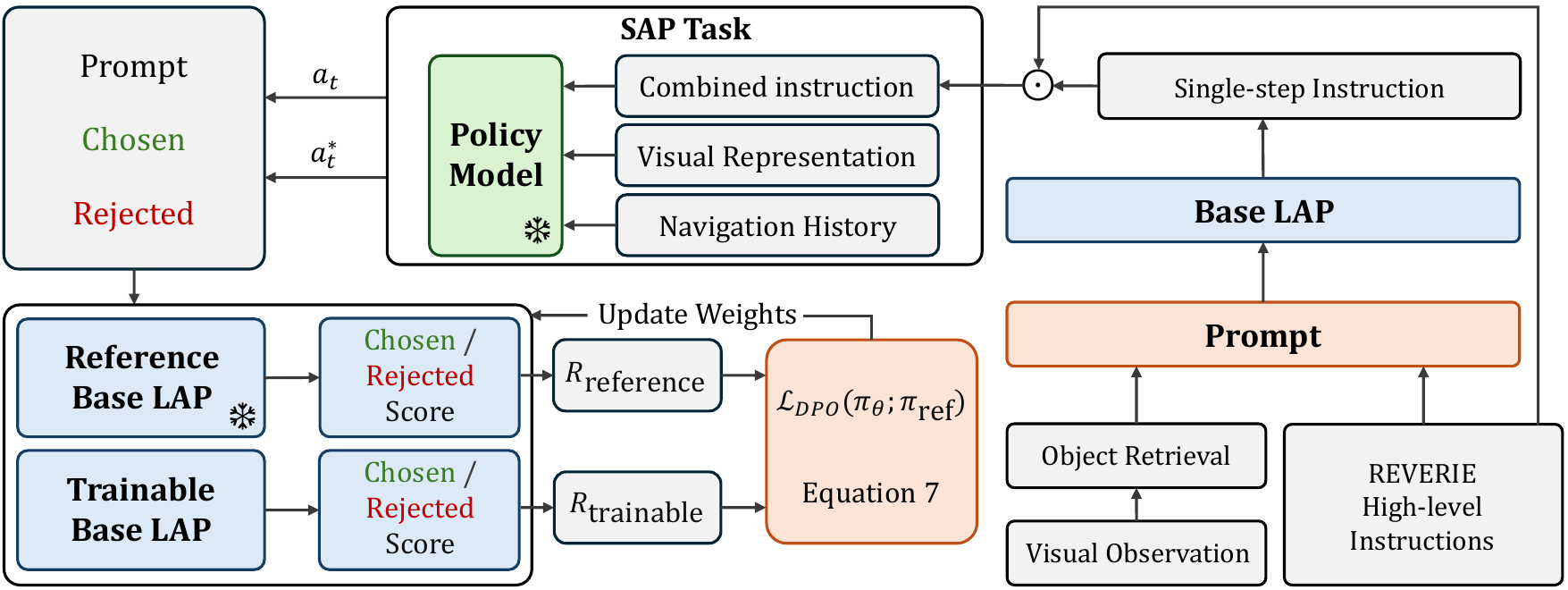}
\end{center}
   \caption{The proposed pipeline for fine-tuning base LAP in DPO setting. LAP first generates a single-step instruction, and then the policy model predicts the action regarding the combined instructions, current visual representation, and the random navigation history. The generated navigation plan by base LAP is selected as chosen if the action leads to the correct prediction, otherwise, it is selected as rejected. The final stage of fine-tuning starts after the preference dataset construction. Note that a frozen copy of LAP is incorporated into the DPO process to control the deviation from the base LAP. {\tiny\img{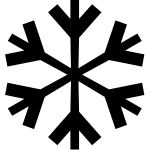}} 
   indicates the model is frozen.
   }
\label{fig::dpo}
\end{figure*}

\subsubsection{Direct Preference Optimization.}
Reinforcement learning from human feedback (RLHF) is now widely established as the final training stage for LLMs to align the output of a model with human expectations. Although effective, this technique is complex and prone to instability. DPO formulation skips the reward modeling and directly optimizes the policy on preference-annotated data. Our main goal for utilizing DPO in the fine-tuning process is to incorporate feedback from the environment and the action prediction process of the policy model. As depicted in Fig.~\ref{fig::dpo}, we propose to fine-tune LAP in three steps: 1) consider a policy model enabling us to distinguish between preferred and dis-preferred single-step instructions, 2) construct the dataset based on the predicted action, and 3) fine-tune the base LAP outputted from the previous stage. To accomplish steps 1 and 2, the SAP task~\cite{chen_2021_history} is employed to collect the correct and wrong single-step instructions. SAP is formulated as a classification task that deploys imitation learning. In this proxy task, the model is asked to predict the next action based on the instruction, the current observation, and history from expert demonstration, \textit{i.e.}, $(\mathcal{V}_1, ..., \mathcal{V}_T)$, where $\mathcal{V}_i$ denotes the $i-th$ viewpoint in the trajectory.

The DPO trainer requires the dataset comprising three key entries: prompt, chosen, and rejected, where the prompt includes the context inputs, chosen contains the sub-instructions leading to selecting the correct viewpoint, and rejected consists of misleading action plans resulting in an inaccurate prediction of the next location. At time step $t$, we construct the prompt based on the high-level instruction $HI_{\mathcal{V}_{1 \rightarrow T}}$ and the current observation and then ask the base LAP to generate a fine-grained language instruction $FI_{v_{t, i}}$. At this moment, the policy model receives visual information, randomly selected navigation history, and the concatenation of high-level and fine-grained instructions, \textit{i.e.}, $HI_{\mathcal{V}_{1 \rightarrow T}} \odot FI_{v_{t, i}}$, to predict the next action $\mathcal{A}_t$. Using the ground-truth action regarding the shortest path in the constructed map in SAP, we realize whether $FI_{v_{t, i}}$ results in the correct action or not. By this approach, we select the generated single-step instruction as chosen or rejected and finalize the dataset for the DPO trainer. The policy formulation of DPO can be defined as:

\begin{equation}
    \max_{\pi_{\theta}} ~ \mathbb{E}_{(x,y_{w},y_{l}) \sim \mathcal{D}} ~
    \log \sigma \left( \beta \log \frac{\pi_{\theta}(y_{w}|x)}{\pi_{\text{ref}}(y_{w}|x)} \right) - \left( \beta \log \frac{\pi_{\theta}(y_{l}|x)}{\pi_{\text{ref}}(y_{l}|x)} \right)
\end{equation}

\vspace{4mm}

\noindent where $x$, $y_w$, and $y_l$ are the prompt, good (chosen), and bad (rejected) responses, respectively. $\pi_\text{ref}$ is the reference model and $\pi_\theta$ is the model we are optimizing. The weights of the reference model are frozen during the training. In fact, the reference model prevents deviating too far from the base LAP model. For the SAP task, we utilize the same pre-trained model of the baseline and keep the model weights frozen to only distinguish between correct and wrong actions.

\section{Experiments}
\label{sec:experiments}

\subsection{Implementation Details}
\label{experiments::implementation_details}
To have a fair comparison, we make no modifications to the baseline implementation and use the same pre-trained model and augmented data in HM3D-DUET~\cite{chen_2022_hm3d}. We merely extend the high-level instruction by adding generated single-step action plans in each location. Moreover, our method is not integrated into the pre-training tasks and is only fine-tuned on the downstream REVERIE task for 20k iterations on a single NVIDIA 3090 GPU. We use AdamW~\cite{loshchilov_2017_decoupled} optimizer, and the learning rate is $10^{-5}$ during the training. Viewpoint images and object bounding boxes are encoded by ViT-B/16~\cite{dosovitskiy_2020_image} pre-trained on ImageNet~\cite{russakovsky_2015_imagenet}, similar to the baseline. To select the top 5 most relevant objects, the CLIP~\cite{radford_2021_Learning} model is utilized. For the LLM, we utilize the publicly available LLAMA 2-CHAT~\cite{touvron_2023_llama} with 7 billion parameters.

\subsection{Dataset}
\label{experiments::dataset}
To evaluate our proposed method, we mainly focus on the challenging REVERIE \cite{qi_2020_reverie} benchmark as the goal-oriented task, which includes concise high-level instructions. Hence, the agent is required to explore the environment efficiently and localize the remote object. REVERIE is the combination of R2R navigation and referring expression tasks in which the agent needs to navigate through the environment to identify the referred object that is not visible in the first view. In this dataset, the average length of instructions is 18 words. Moreover, more than 4,000 target objects fall into 489 categories. Expert paths vary in length from 4 to 7 steps.

\subsection{Evaluation Metrics}
\label{experiments::evaluation_metrics}
We utilize the widely used and standard metrics for performance evaluation on REVERIE. \textbf{success rate (SR):} the ratio of successful tasks, meaning the stopping point to the target location is less than 3 meters, \textbf{success rate penalized by path length (SPL):} the trade-off between SR and trajectory length indicating the navigation efficiency. REVERIE also defines \textbf{remote grounding success rate (RGS):} the proportion of localizing the correct object at the stopping point, \textbf{remote grounding success rate weighted by path length (RGSPL):} the RGS which is penalized by the trajectory length. RGS and RGSPL correspond to the object grounding ability of the agent, while the rest are used to evaluate the navigation performance. Metrics that are penalized by path length are the most important measures, \textit{i.e.}, SPL and RGSPL.

\begin{table*}[t!]
  \Huge
  \caption{Comparison of the agent performance with state-of-the-art methods on the REVERIE dataset.}
  \begin{center}
  \resizebox{0.97\textwidth}{!}{
  \begin{tabular}{l|cc|cc|cc|cc}
    \hline
    \multicolumn{1}{c|}{\multirow{3}{*}{Methods}} & \multicolumn{4}{c|}{Validation Unseen} & \multicolumn{4}{c}{Test Unseen} \Tstrut \\
    \cline{2-9} & \multicolumn{2}{c|}{Navigation} & \multicolumn{2}{c|}{Grounding} & \multicolumn{2}{c|}{Navigation} & \multicolumn{2}{c}{Grounding}  \\
    \cline{2-9} & 
    \multicolumn{1}{c}{SR$\uparrow$} & \multicolumn{1}{c|}{SPL$\uparrow$} & \multicolumn{1}{c}{RGS$\uparrow$} & \multicolumn{1}{c|}{RGSPL$\uparrow$} & \multicolumn{1}{c}{SR$\uparrow$} & \multicolumn{1}{c|}{SPL$\uparrow$} & \multicolumn{1}{c}{RGS$\uparrow$} & \multicolumn{1}{c}{RGSPL$\uparrow$} \Tstrut\\
    \hline

    Human & - & - & - & - & 81.51 & 53.66 & 77.84 & 51.44  \\

    \hline

    CKR~\cite{gao_2021_room} & 19.14 & 11.84 & 11.45 & - & 22.00 & 14.25 & 11.60 & - \\

    SIA~\cite{lin_2021_scene} & 31.53 & 16.28 & 22.41 & 11.56 & 30.80 & 14.85 & 19.02 & 9.20  \\

    ORIST~\cite{qi_2021_road} & 16.84 & 15.14 & 8.52 & 7.58 & 22.19 & 18.97 & 10.68 & 9.28 \\

    HAMT~\cite{shizhe_2021_hamt} & 32.95 & 30.20 & 18.92 & 17.28 & 30.40 & 26.67 & 14.88 & 13.08  \\

    Airbert~\cite{guhur_2021_airbert} & 27.89 & 21.88 & 18.23 & 14.18 & 30.28 & 23.61 & 16.83 & 13.28  \\

    RecBERT~\cite{hong_2021_recbert} & 30.67 & 24.90 & 18.77 & 15.27 & 29.61 & 23.99 & 16.50 & 13.51  \\

    HOP~\cite{qiao_2022_hop} & 31.78 & 26.11 & 18.85 & 15.73 & 30.17 & 24.34 & 17.69 & 14.34  \\

    TD-STP~\cite{zhao_2022_target} & 34.88 & 27.32 & 21.16 & 16.56 & 35.89 & 27.51 & 19.88 & 15.40 \\
    
    Lily~\cite{lin_2023_lily} & 48.11 & 34.43 & 32.15 & 23.43 & 54.32 & 37.34 & 32.02 & 21.94 \\

    DUET~\cite{chen_2022_duet} & 46.98 & 33.73 & 32.15 & 23.03 & 52.51 & 36.06 & 31.88 & 22.06  \\

    BEVBert~\cite{an_2023_bevbert} & 51.78 & 36.37 & 34.71 & 24.44 & 52.81 & 36.41 & 32.06 & 22.09 \\

    BSG~\cite{liu_2023_bsg} & 52.12 & 35.59 & 35.36 & 24.24 & 56.45 & 38.70 & 33.15 & 22.34 \\

    AZHP~\cite{gao_2023_azhp} & 48.31 & 36.63 & 34.00 & 25.79 & 51.57 & 35.85 & 32.25 & 22.44  \\
        
    ACK~\cite{mohammadi_2024_augmented} & 47.49 & 34.44 & 32.66 & 23.92 & 53.97 & 37.89 & 32.77 & 23.15  \\

    VER~\cite{liu_2024_volumetric} & 55.98 & 39.66 & 33.71 & 23.70 & 56.82 & 38.76 & 33.88 & 23.19 \\

    KERM~\cite{li_2023_kerm} & 50.44 & 35.38 & 34.51 & 24.45 & 52.43 & 39.21 & 32.39 & 23.64  \\

    \hline

    HM3D-DUET~\cite{chen_2022_hm3d} & 55.89 & 40.85 & 36.58 & 26.76 & 55.17 & 38.88 & 32.23 & 22.68  \\

    PEAP-LLM (Ours) & \textbf{58.65} & \textbf{44.85} & \textbf{39.14} & \textbf{29.96} & \textbf{56.01} & \textbf{40.98} & \textbf{33.85} & \textbf{24.88} \\

    \hline
    
  \end{tabular}}
\end{center}
\label{tab:reverie_results}
\end{table*}

\subsection{Comparison to State-of-the-Arts}
\label{experiments::comparison}
We thoroughly compare our proposed method with state-of-the-art approaches in Table~\ref{tab:reverie_results} using the most important REVERIE data splits, \textit{i.e.}, validation unseen and test unseen, which reflects the generalization ability and performance of the agent in unseen environments. To provide a fair comparison with the baseline approach, we use the same pre-trained model with no changes in implementation. The reported results in this table correspond to using LoRA for the SFT, followed by the DPO trainer. As shown in Table~\ref{tab:reverie_results}, we outperform previous state-of-the-art methods in all metrics on both validation unseen and test unseen splits. In particular, our proposed method outperforms HM3D-DUET by a large margin of $4.00$\% on SPL and $3.20$\% on RGSPL for the validation unseen split and $2.10$\% on SPL and $2.20$\% on RGSPL for the test unseen split. Note that SPL and RGSPL are the main metrics that represent the effectiveness of a method for navigation and object grounding performance. These promising results indicate that PEAP-LLM has great potential to significantly enhance the general performance of the embodied agent.

\begin{table}[t!]
  \tiny
  \caption{Ablation of utilizing the LGP individually along with its integration with LAP in our proposed approach on the REVERIE validation unseen split.}
  \begin{center}
  \resizebox{0.80\columnwidth}{!}{\def\arraystretch{1}
  \begin{tabular}{c|ccccc}
    \hline
    
    \multicolumn{1}{c|}{\multirow{2}{*}{Modules}} & \multicolumn{2}{c}{Navigation} & \multicolumn{2}{c}{Grounding} \Tstrut \\
    
    & SR$\uparrow$ & SPL$\uparrow$ & RGS$\uparrow$ & RGSPL$\uparrow$ \Tstrut \\
    
    \hline

    Baseline & 54.36 & 40.58 & 36.03 & 26.81 \\

    LGP & 55.47 & 42.04 & 36.95 & 27.77 \\

    LAP + LGP & \textbf{58.65} & \textbf{44.85} & \textbf{39.14} & \textbf{29.96} \\
 
    \hline
  \end{tabular}}
\end{center}
\label{tab::ablation_study::modules}
\vspace{-4mm}
\end{table}

\begin{table}[t!]
\caption{Ablation of employing different approaches for fine-tuning the LLM separately or together in our proposed method on the REVERIE validation unseen split.}
  \begin{center}
  \resizebox{0.85\columnwidth}{!}{
  \def\arraystretch{1}
  \begin{tabular}{cccc|cccc}
    \hline

    \multicolumn{4}{c|}{Fine-tuning Approaches} & \multicolumn{2}{c}{Navigation} & \multicolumn{2}{c}{Grounding} \Tstrut \\
    
    \multicolumn{1}{c}{BLLM} & \multicolumn{1}{c}{PT} & \multicolumn{1}{c}{LoRA} & \multicolumn{1}{c|}{DPO} & SR$\uparrow$ & SPL$\uparrow$ & RGS$\uparrow$ & RGSPL$\uparrow$ \Tstrut \\
    
    \hline

    $\checkmark$ & $\times$ & $\times$ & $\times$ & 55.52 & 42.45 & 37.63 & 28.52 \\
    
    $\times$ & $\checkmark$ & $\times$ & $\times$ & 56.55 & 43.08 & 37.94 & 28.93  \\

    $\times$ & $\times$ & $\checkmark$ & $\times$ & 57.34 & 44.25 & 38.02 & 29.18 \\

    $\times$ & $\checkmark$ & $\times$ & $\checkmark$ & 57.60 & 44.32 & 38.65 & 29.35 \\

    $\times$ & $\times$ & $\checkmark$ & $\checkmark$ & \textbf{58.65} & \textbf{44.85} & \textbf{39.14} & \textbf{29.96} \\

    \hline
  \end{tabular}}
\end{center}
\label{tab::ablation_study::approaches}
\vspace{-4mm}
\end{table}

\subsection{Ablation Study}
\label{experiments::ablation_study}
The contribution of each module is assessed through comprehensive experiments shown in Table~\ref{tab::ablation_study::modules}. Furthermore, the effectiveness of each approach for LLM fine-tuning is presented in Table~\ref{tab::ablation_study::approaches}. Our proposed method is ablated on the validation unseen split of REVERIE.

\vspace{1mm}
\noindent \textbf{Contribution of Different Modules.}
We evaluate the effect of each module on the navigation and object grounding performance of the agent. LGP shows the results for merely using the LLM goal planner, which identifies the target object and room. LGP + LAP demonstrates the performance of PEAP-LLM when the two proposed modules are integrated to generate single-step instructions. As Table~\ref{tab::ablation_study::modules} suggests, even leveraging merely the goal-oriented plan can improve the baseline in all metrics, which confirms the benefit of LGP. Moreover, by adding LAP as the path planner, performance is further enhanced by a noticeable margin, showing the effectiveness of our novel approach for action planning and LLM fine-tuning on the REVERIE downstream task.

\vspace{1mm}
\noindent \textbf{Impact of Different Approaches for Fine-tuning.}
The performance of our proposed parameter-efficient action planner is assessed in different scenarios, such as utilizing the base LLM and its fine-tuned version using different PEFT techniques, including prefix tuning and LoRA, along with their combination with the DPO trainer. BLLM, PT, LoRA, and DPO represent the results for the base LLM, prefix tuning, low-rank adaptation, and direct preference optimization, respectively. We evaluate the performance of our method using prefix tuning and LoRA individually, along with their integration with DPO, which is usually established as the final fine-tuning stage. According to reported results in Table~\ref{tab::ablation_study::approaches}, LoRA is more powerful than prefix tuning for single-step action prediction. When the supervised fine-tuned model employs the DPO technique for fine-tuning the base LAP regarding the feedback from the environment, the performance of the agent is further improved in all metrics. The best results are associated with exploiting LoRA for SFT, followed by the DPO trainer as the final stage.

\begin{figure*}[t!]
\begin{center}
\includegraphics[width=0.99\linewidth]{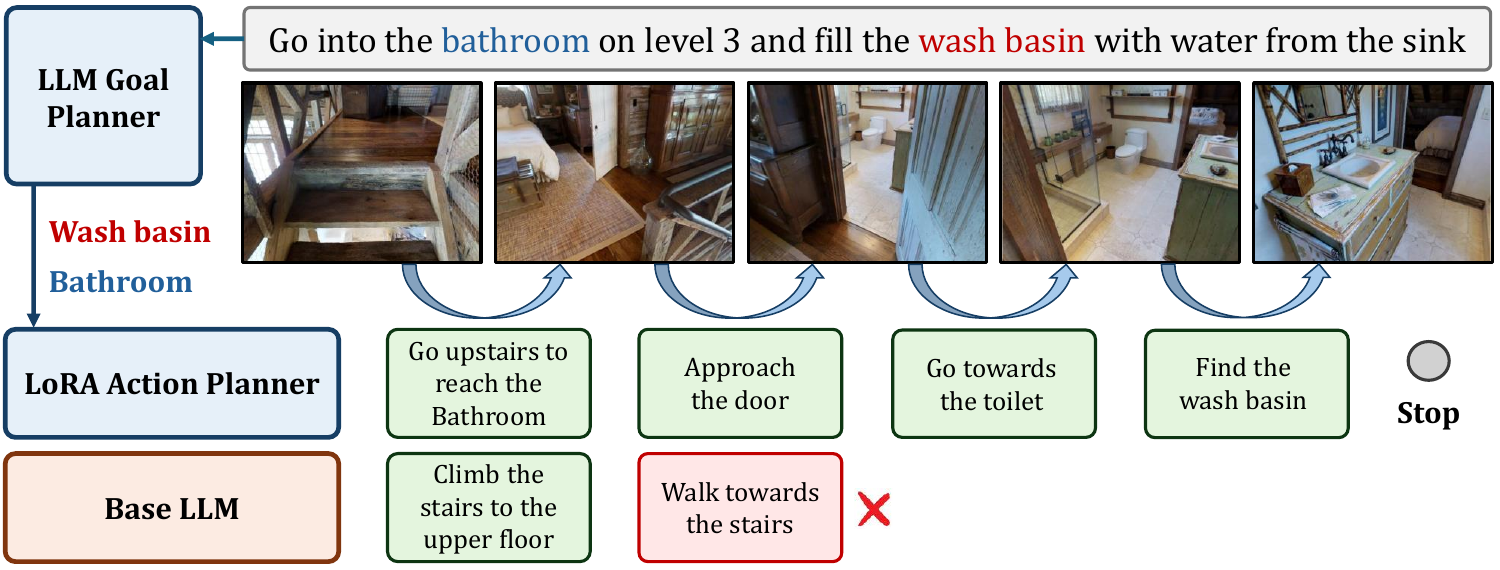}
\end{center}
   \caption{Visualization example of navigation performance for comparing LAP and the base LLM. At first, LGP extracts ``\textit{wash basin}'' and ``\textit{bathroom}''. Afterward, both base LLM and LAP generate sub-instructions using the goal-oriented plan, REVERIE instruction, and visual observation as input. The agent interacts with LAP on the fly to obtain a single-step action plan in each viewpoint to find the correct navigable direction. However, the base LLM fails to provide effective sub-instructions.}
\label{fig::qualitative}
\end{figure*}

\begin{table}[t!]
  \Huge
  \caption{Impact of defining the object list and the LLM fine-tuning on using the visual landmarks correctly.}
  \vspace{-1mm}
  \begin{center}
  \resizebox{0.85\columnwidth}{!}{\def\arraystretch{1.1}
  \begin{tabular}{cc|c|c}
    \hline
    
    \multicolumn{2}{c|}{Fine-tuning} & \multicolumn{1}{c|}{\multirow{2}{*}{Object List}} & \multicolumn{1}{c}{\multirow{2}{*}{Correct Usage of Visual Landmarks ($\%$)}} \Tstrut \\
    
    Prompt Tuning & LoRA &  &  \Tstrut \\
    
    \hline
    
    $\times$ & $\times$ & $\times$ & 7.37 \\

    $\times$ & $\times$ & $\checkmark$ & 96.47 \\

    $\checkmark$ & $\times$ & $\checkmark$ & 97.89 \\

    $\times$ & $\checkmark$ & $\checkmark$ & 98.96 \\

    \hline
  \end{tabular}}
\end{center}
\label{tab::quantitative::hallucination}
\end{table}

\pgfplotstableread[row sep=\\,col sep=&]{
    interval & Base LLM & Fine-tuned LLM \\
    shelf     & 6343 & 6897  \\
    fireplace    & 724 & 4884 \\
    stairs   & 18348 & 8524 \\
    mirror   & 15764  & 5698 \\
    step      & 1373  & 3729 \\
    rug      & 21250  & 9163 \\
    cabinet      & 23279  & 8389 \\
    desk      & 138  & 2696 \\
    chair      & 1454  & 5568 \\
    table      & 6489  & 3075 \\
    case      & 281  & 6647 \\
    fridge      & 6774  & 4901 \\
    lamp      & 8030  & 2722 \\
    sofa      & 1217  & 3567 \\
    }\mydata

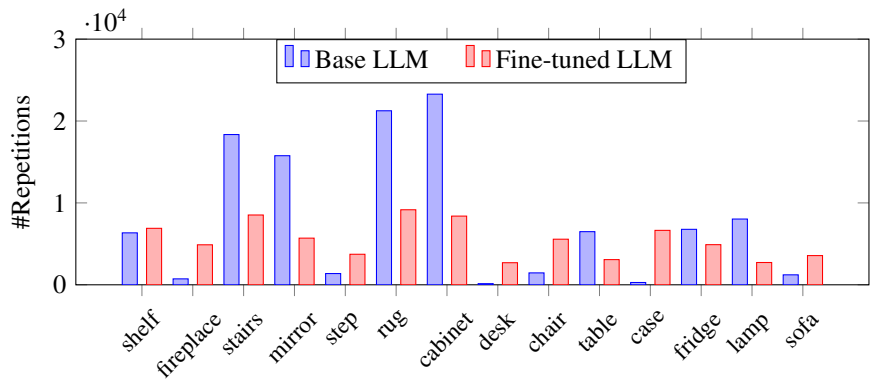
\begin {figure*}[!hbtp]
\centering
\begin{tikzpicture}
    \begin{axis}[
            ybar=3.5pt,
            bar width=.2cm,
            width=\textwidth,
            height=.4\textwidth,
            legend style={at={(0.77,1)},
                anchor=north east,legend columns=-1},
            /tikz/every even column/.append style={column sep=1.2em},
            symbolic x coords={shelf, fireplace, stairs, mirror, step, rug, cabinet, desk, chair, table, case, fridge, lamp, sofa},
            xtick=data,
            nodes near coords style={anchor=east,rotate=90,xshift=27pt},
            ymin=0,ymax=30000,
            ylabel={\#Repetitions},
            xticklabel style={font=\small, xshift=-1pt, rotate=45}]
        ]
        \addplot table[x=interval,y=Base LLM]{\mydata};
        \addplot table[x=interval,y=Fine-tuned LLM]{\mydata};
        \legend{Base LLM, Fine-tuned LLM}
    \end{axis}
\end{tikzpicture}
\caption{Number of repetitions for most common objects in generated single-step instructions for base and fine-tuned LLMs.}
\label{fig::object_repetition}
\end{figure*}

\subsection{Qualitative Analysis}
In this section, we use an example of the validation unseen split in REVERIE to visualize the functions of the proposed modules. Note that the baseline agent fails to reach the destination and localize the target object. As illustrated in Fig.~\ref{fig::qualitative}, the baseline agent can interact with LAP on the fly as the path planner to obtain the single-step action plan in each step and select the correct navigable direction. However, the base LLM fails to provide the correct sub-instruction in the second viewpoint, which leads to an incorrect location. This example shows how our proposed approach for LLM fine-tuning helps the agent explore the environment more efficiently.

\subsection{Quantitative Analysis}
In general, evaluating the hallucination generation is a complicated and challenging task. In addition, generating hallucinations is unavoidable when using LLMs. Hence, we investigate the effect of constraining the output by designing an appropriate prompt and LLM fine-tuning to mitigate this problem. Table~\ref{tab::quantitative::hallucination} demonstrates how defining an appropriate object list in the prompt helps generate sub-instructions with accurate visual landmarks. Additionally, we examine the impact of fine-tuning on producing correct data. This table implicitly shows the effectiveness of our two-step fine-tuning approach to mitigate hallucination generation. Note that this table only shows the usage of objects that are truly present in the scene, not the performance of the agent. Also, we investigate how LLM fine-tuning prevents generating biased information. As illustrated in Fig.~\ref{fig::object_repetition}, LLM fine-tuning results in a more balanced use of visual clues in generated single-step instructions, whereas the base LLM is biased to use certain objects over others.

\section{Conclusion}
\label{sec:conclusion}
In this paper, we propose a parameter-efficient action planner using large language models (PEAP-LLM) that enables the agent to obtain a single-step navigation plan in each location interactively. In particular, our model consists of two modules, LLM goal planner (LGP) and LoRA action planner (LAP). The former module is responsible for extracting the goal of REVERIE instructions, including the target object and target room. The latter module obtains the goal-oriented plan, perceives the environment, and receives the high-level instruction to generate a single-step action plan. Finally, the baseline policy model predicts the next action using combined high-level and generated instructions. To improve the LLM performance for the navigation task and mitigate generating hallucinations and biased data, we present a two-step approach for LLM fine-tuning, including supervised fine-tuning (SFT) and direct preference optimization (DPO). We have conducted extensive experiments on REVERIE, and the promising results show the effectiveness and generalization ability of our proposed method.

\vspace{1mm}
\noindent\textbf{Limitations and Future Work.} Generating a single-step instruction at each location may impose computational overhead. Therefore, defining a metric to quantify the confidence level of the agent can mitigate the overhead so that the agent asks LAP for an action plan only when the uncertainty is detected. Additionally, this approach allows the agent to incorporate additional contextual information into the prompt, such as the relative orientation of visible objects with respect to the angle and elevation of the agent, leading to the generation of more accurate single-step navigation plans.

\bibliographystyle{elsarticle-num}
\bibliography{elsarticle-template-num}

\end{document}